\documentclass[runningheads]{llncs}
\usepackage[T1]{fontenc}
\usepackage{graphicx}
\usepackage{amsmath,amsfonts,amssymb} 
\usepackage{booktabs}
\usepackage{subcaption}
\usepackage{xcolor}
\begin{document}
\title{Physics-informed Discovery of State Variables in Second-Order and Hamiltonian Systems}
\author{Félix Chavelli\inst{1} \and
Zi-Yu Khoo\inst{1} \and Dawen Wu\inst{2}\inst{1} \and  Jonathan Sze Choong Low\inst{3} \and 
Stéphane Bressan\inst{1}\inst{3}}
\authorrunning{Chavelli et al.}
\titlerunning{Physics-informed Discovery of State Variables}
\institute{National University of Singapore. 21 Lower Kent Ridge Rd, Singapore 119077
\email{{chavelli,khoozy}@comp.nus.edu.sg, steph@nus.edu.sg}\\
\and CNRS@CREATE LTD, 1 Create Way, Singapore 138602  \email{dawen.wu@cnrsatcreate.sg} \\
\and Singapore Institute of Manufacturing Technology, Agency for Science, Technology and Research (A*STAR), Singapore 138634 \email{sclow@simtech.a-star.edu.sg}
\\
}
\maketitle              % typeset the header of the contribution
\begin{abstract}

The modeling of dynamical systems is a pervasive concern for not only describing but also predicting and controlling natural phenomena and engineered systems. Current data-driven approaches often assume prior knowledge of the relevant state variables or result in overparameterized state spaces.  Boyuan Chen and his co-authors proposed a neural network model that estimates the degrees of freedom and attempts to discover the state variables of a dynamical system. Despite its innovative approach, this baseline model lacks a connection to the physical principles governing the systems it analyzes, leading to unreliable state variables.
This research proposes a method that leverages the physical characteristics of second-order Hamiltonian systems to constrain the baseline model. The proposed model outperforms the baseline model in identifying a minimal set of non-redundant and interpretable state variables.

\end{abstract}
\section{Introduction}

Dynamical systems, systems that change over time, pervade the natural and engineered world, embodying the complex interactions and evolution observed across a multitude of fields, from physics and biology to economics and engineering. Their universality and applicability in modeling real-world phenomena underscore the imperative for their study, motivating works that offer insights into system behavior, prediction, and control~\cite{Wang2023PredRNN,Gao_2022_CVPR,wu2024earthfarseer,Sasaki19,gao2022bayesian}. 
However, most data-driven methods for modeling dynamical systems still rely on the assumption that the relevant state variables are already known~\cite{chen2022discovery,gao2022bayesian} or use more parameters than necessary to represent possible configurations of the state variables~\cite{wu2024earthfarseer,Gao_2022_CVPR}.

Recent work by Chen et al. \cite{chen2022discovery} proposed a neural network model for dynamical system analysis that identifies the relevant state variables. The model estimated the degrees of freedom and discovered the state variables of a dynamical system from its images using an encoder-decoder architecture~\cite{chen2022discovery}. Chen et al.'s model marks a significant stride toward understanding system dynamics. The identification of state variables is highly synergistic with the new paradigm of physics-informed machine learning~\cite{Karniadakis2021} as it enables the incorporation of fundamental physics principles in data-driven machine learning models through the identified state variables. 

%%%%%%%%%%%
%and , which synergizes the generalization capabilities of neural networks with the rigor of physical laws \cite{raissi2019physics} by systematically incorporating various physical information or constraints into machine learning models. 

%Observational biases are introduced directly through data that embody the underlying physics, or carefully crafted data augmentation procedures. With sufficient data to cover the input domain of a regression task, machine learning methods have demonstrated remarkable power in achieving accurate interpolation between the dots~\cite{Karniadakis2021}. 
%Learning biases are soft constraints introduced by appropriate loss functions, constraints, and inference algorithms that modulate the training phase of a machine learning model to explicitly favor convergence towards solutions that adhere to the underlying physics~\cite{Karniadakis2021}. 
%Inductive biases are prior assumptions incorporated by tailored interventions to a machine learning model architecture, so regressions are guaranteed to implicitly and strictly satisfy a set of given physical laws~\cite{Karniadakis2021}. 

Yet, Chen et al.'s proposed model is detached from the physical underpinnings of the systems it models~\cite{chen2022discovery}; the proposed model uses an external intrinsic dimension estimator~\cite{Bellman1961,levina2004maximum,10.1007/978-3-031-39847-6_18} which is ignorant of physics principles to estimate a non-integer number of degrees of freedom. Furthermore, the obtained state variables are correlated, possibly redundant or entangled, with no guarantee of interpretability. %We therefore turn to physics-informed symbolic regression methods leverage physics concepts such as parsimony~\cite{Udrescu2020,brunton2016}, invariances between the generalized coordinates~\cite{Jiang_Xue_2024} and in the governing equation~\cite{Udrescu2020,Xing_Salleb-Aouissi_Verma_2021}, or smartly sample the space of generalized coordinates~\cite{jin2023} to discover the governing equations that PINNs model.
Therefore, this research pivots towards harnessing the physical characteristics of dynamical systems, specifically their second-order~\cite{Ginsberg2008} and Hamiltonian~\cite{Meyer1992,Nolte2018} characteristics. By leveraging physics-informed machine learning, we propose to incorporate physics knowledge into the machine learning model: observational biases are introduced directly through data that embody the underlying physics~\cite{Karniadakis2021}, learning biases are introduced through appropriate loss functions which modulate convergence towards solutions that adhere to the underlying physics~\cite{Karniadakis2021,bertalan_2019}, and inductive biases are incorporated by tailored interventions to a machine learning model architecture~\cite{Karniadakis2021}. 
These embed physical constraints into Chen et al.'s baseline model, to identify a minimal set of non-redundant, interpretable state variables. %adhering to conventional coordinates to ensure the model's comprehensibility and interoperability. 
% Our central research question seeks to exploit these physical properties to achieve an optimal representation of the system's dynamics, offering a balance between simplicity and accuracy that improves existing methodologies.

We provide a brief background of dynamical systems and autoencoders in Section~\ref{sec:background}. We then introduce several modifications to Chen et al.'s model using physics-informed machine learning in Section~\ref{sec:method}. In Section~\ref{sec:exp}, we evaluate, compare, and find that the proposed model outperforms Chen et al.'s model on their original dataset by accurately identifying the number of state variables. Additionally, the modifications merge and simplify Chen et al.'s model and improve the interpretability of the identified state variables.

\section{Background}~\label{sec:background}
Dynamical systems are defined by a set of time-dependent equations that characterize the evolution of a system's state over time~\cite{Ginsberg2008}, and are %These systems are specified by state variables and the differential equations that describe how these variables evolve. Typically, 
% A dynamical system is
represented as
\begin{equation}
    \dot{\chi} = f(\chi, t)
\end{equation}
where $\chi$ is a vector of the state variables of length equivalent to the number of degrees of freedom of the dynamical system, $\dot{\chi}$ is the time derivative of $\chi$ %the derivative of the state vector $x$ 
with respect to time $t$, and $f$ is a function that defines the system dynamics \cite{Ginsberg2008}. 
%The state space, or phase space, encompasses all possible states of the system, with its dimension equal to the number of state variables required for a complete description of the system . 

Autonomous second-order dynamical systems are a class of dynamical systems characterized by an even number of state variables where half of them describe the position ($\tau$), and the other half represent the associated momentum ($\rho$) \cite{Ginsberg2008}. 
Hamiltonian dynamical systems are a class of dynamical systems whose dynamics are governed by Hamilton's equations
\begin{equation}
    \dot{\tau} = \frac{\partial H}{\partial \rho}, \quad \dot{\rho} = -\frac{\partial H}{\partial \tau}
\end{equation}
where $H$ represents the total energy or Hamiltonian of the dynamical system \cite{Meyer1992,Nolte2018}, and is a conserved quantity. Hamiltonian neural networks~\cite{bertalan_2019,Karniadakis2021,Greydanus_hamiltoniannn_2019} are physics-informed machine learning models that incorporate learning biases based on Hamilton's equations within neural networks. They regress the Hamiltonian of a dynamical system directly from its state variables using Hamilton's equations and enforce the invariance of the total energy of the dynamical system. 

% , combining kinetic energy $T$ with potential energy $V$. The dynamics of Hamiltonian systems are governed by Hamilton's equations:$$\dot{q} = \frac{\partial H}{\partial p}, \quad \dot{p} = -\frac{\partial H}{\partial q}$$ which describe the time evolution of the generalized coordinates and velocities, highlighting the system's energy conservation . A recent physics-informed machine learning model that leverages the Hamiltonian property of dynamical systems is Hamiltonian neural networks~\cite{bertalan_2019,Karniadakis2021,Greydanus_hamiltoniannn_2019}. They incorporate learning biases based on Hamilton's equations, with no prior knowledge of the system's Hamiltonian. These networks are designed to unsupervisedly regress both the vector field and the Hamiltonian of a dynamical system directly from discrete observations of its state space or vector field \cite{WainwrightCalculus2020,Sayama_2015}. This approach offers a robust framework for analyzing dynamical systems, bridging the gap between theoretical physics and data-driven machine learning models.

% In engineering dynamics, autonomous second-order systems are prevalent, characterized by time-independent governing equations and 

An autoencoder is a neural network comprising an encoder function, which constructs an encoding of the input, and a decoder function, which produces a reconstruction of the input~\cite{GoodBengCour16}. Sandwiched between them is a hidden layer that describes a code, or latent variables, used to represent the input~\cite{GoodBengCour16}. Autoencoders are designed to be unable to learn to copy perfectly, usually by limiting the number of latent variables~\cite{GoodBengCour16}. 

% Autoencoders are a class of neural networks designed for unsupervised learning of compressed, efficient representations of input data. They work by encoding input data into a latent (hidden) space representation and then reconstructing the input from this representation. The loss function of such models consists of a single reconstruction term (e.g., mean squared error between the input and its reconstruction).

Variational Autoencoders (VAEs) extend the autoencoder framework by introducing a probabilistic approach to encoding inputs~\cite{Kingma2013}. VAEs modify the loss function of an autoencoder by adding a Kullback-Leibler (KL) divergence term in the loss function,
\begin{equation}
\mathcal{L}_{VAE} = -\beta \cdot \mathbb{E}_{q(z|x)}[\log p(x|z)] +  KL(q(z|x) \| p(z))
\end{equation}
where $x$ is the input and $z$ its latent representation, $-\mathbb{E}_{q(z|x)}[\log p(x|z)]$ is the reconstruction loss in a variational context, \( q(z|x) \) denotes the encoder's distribution and \( p(z) \) denotes a prior distribution \( p(z) \). $\beta$ adjusts the respective weights of the two terms. It can be placed on either the reconstruction or KL divergence term. The KL divergence term penalizes deviations of the learned distribution in the latent space from a chosen prior Gaussian distribution.

The KL divergence term enforces independence among the latent variables by pushing the encoded distributions to resemble the prior~\cite{asperti2019sparsity}. Therefore variables that do not contribute significantly to reducing the reconstruction loss become redundant and converge to the non-informative zero prior~\cite{asperti2019sparsity}. This process inherently minimizes the size of the latent space to the number of degrees of freedom of the dynamical system, by eliminating excess variables.
Furthermore, the inclusion of the KL divergence encourages the VAE to find disentangled, semantically meaningful, statistically independent, and causal factors of variation in data~\cite{Kingma2013}. The result is a more interpretable and minimal representation, which facilitates understanding of the underlying structure of the data.

\section{Methodology}~\label{sec:method}
\begin{figure}[h]
    \centering
    \includegraphics[width=1\textwidth]{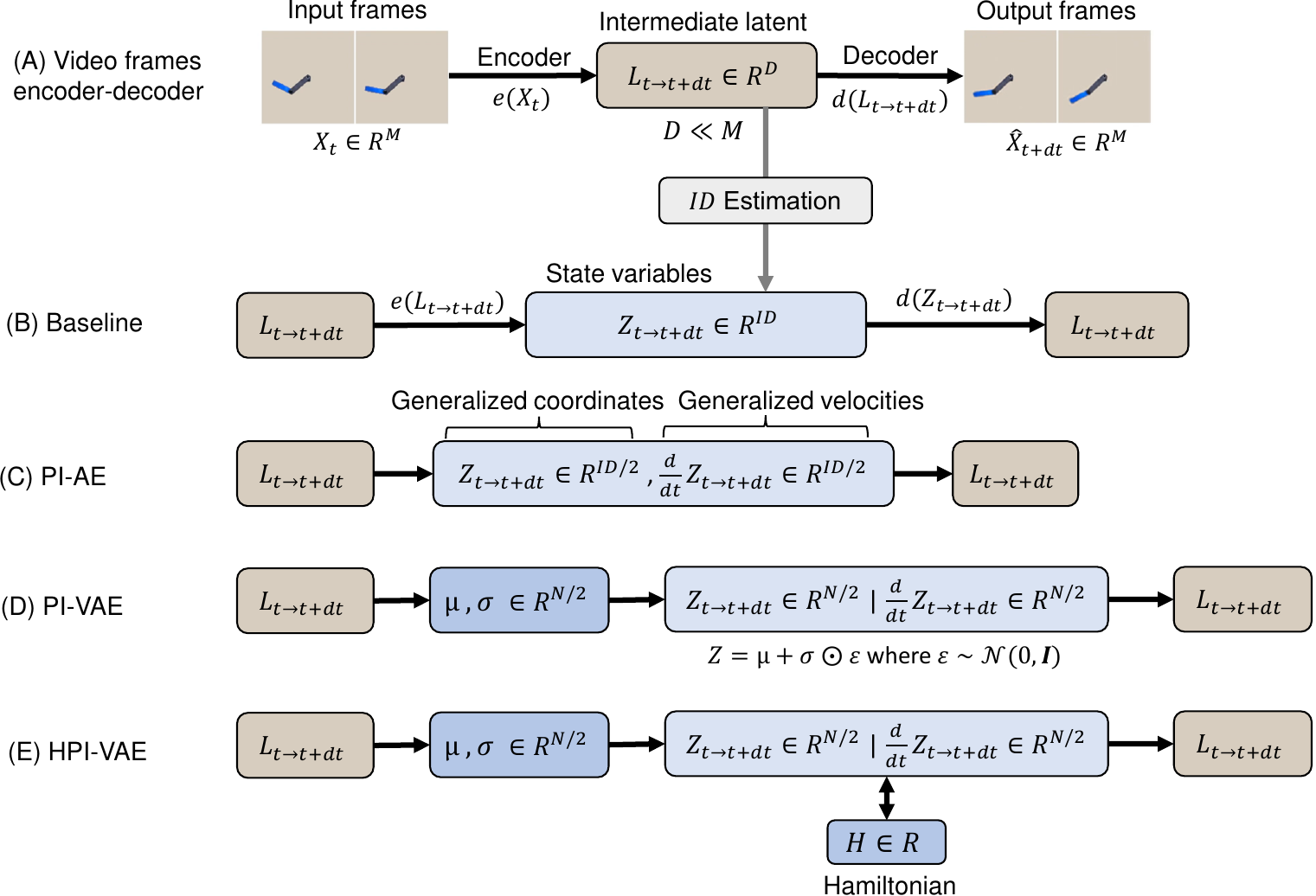}
    \caption{Chen et al.'s baseline (A and B) and our proposed models (C, D, E). }
    \label{fig:diagrams}
\end{figure}
Chen et al.'s baseline model is a nested autoencoder~\cite{chen2022discovery} (Figure~\ref{fig:diagrams}, A and B). In the first step, two consecutive video frames from a dynamical system are input to the outermost autoencoder. A compact representation of the input is extracted and the subsequent two consecutive frames are predicted (Figure~\ref{fig:diagrams}, A). Then, an intrinsic dimension (ID) estimation~\cite{levina2004maximum} of the compact representation is performed. In the second step, the innermost autoencoder further compresses the compact representation into a latent space of dimension ID (Figure~\ref{fig:diagrams}, B). These are the state variables of the dynamical system. %The innermost autoencoder ensures the reconstruction of the compact representation from the outermost autoencoder, to maintain the prediction of the subsequent two frames.

We leverage Karniadakis et al.'s framework to enhance Chen et al.'s model by introducing three proposed models, each incorporating different physics-informed machine learning biases. The three proposed models shown in Figure 1 C, D, and E utilize the outermost autoencoder from Chen et al. (Figure~\ref{fig:diagrams}, A) and make modifications to the innermost autoencoder (Figure~\ref{fig:diagrams}, B). %This approach allows us to explore the impact of integrating different physical principles into the predictive modeling of dynamical systems, highlighting the potential for more accurate and understandable data-driven methods.

\subsection{Observational bias}
The Physics-Informed AutoEncoder (PI-AE) builds on Chen et al.'s baseline. It incorporates an observational bias by enforcing the system's second-order constraint to the latent space of the innermost autoencoder. Following the nomenclature of second-order dynamical systems, pairs of latent variables of the autoencoder are constrained, such that the first represents the position, and the second, the momentum of the dynamical system. %$2n$ dimensions, comprising $n$ generalized positions and their $n$ time derivatives or generalized velocities, where $2n$ = ID is determined by the intrinsic dimension estimator. 

\subsection{Learning bias}
The Physics-Informed Variational AutoEncoder (PI-VAE) builds on the PI-AE and incorporates a learning bias regarding the time-continuity of the dynamical system and the independence of the state variables. The constraint is incorporated by a variational autoencoder. The KL divergence term in the loss function enforces latent sparsity, as the learned distribution in the latent space has a Standard Multivariate Normal Distribution prior. 

% This pushes independence between the latent variables and continuity with respect to the input. The number of state variables of the dynamical system is discovered by the variational autoencoder during training, which renders the external intrinsic dimension estimator redundant. For this purpose, the latent size of the variational autoencoder is set to an arbitrary number $N > 2n$. The choice of this value is discussed in the next section. This Physics-Informed Variational AutoEncoder model is referred to as PI-VAE.

\subsection{Inductive bias}
The Hamiltonian Physics-Informed Variational AutoEncoder (HPI-VAE) builds on the PI-VAE and incorporates an inductive bias. It modifies the PI-VAE architecture to include a Hamiltonian neural network which takes as input the latent variables. %The invariance of the total energy of the dynamical system can be enforced by placing a Hamiltonian neural network on top of the latent space, that takes as input the generalized coordinates and velocities and outputs the Hamiltonian or total energy of the system. 
This model has three terms in its loss function. They are the reconstruction loss, the KL divergence, and Hamilton's equations. 

Table~\ref{tab:modelsummary} sums up the biases introduced in each model. The three modifications are progressively incorporated into the baseline model using observational, learning, and inductive bias respectively. Through a systematic comparison in Section~\ref{sec:exp}, we evaluate the capability of each model to capture the dynamics of various systems in a minimal, interpretable set of latent variables. 

\begin{table}[h]
    \centering
    \caption{Summary of baseline and proposed models}
    \begin{tabular}{l@{\hspace{1em}}c@{\hspace{1em}}c@{\hspace{1em}}c@{}}
    \toprule
    Model & Observational bias & Learning bias & Inductive bias \\
    \midrule
    Baseline & - & - & - \\
    PI-AE & $2^{nd}$ order constraint & - & -\\
    PI-VAE & $2^{nd}$ order constraint & Latent sparsity & - \\
    HPI-VAE & $2^{nd}$ order constraint & Latent sparsity & Hamiltonian conservation\\
    \bottomrule
    \end{tabular}
    \label{tab:modelsummary}
\end{table}

\section{Experiments}
\label{sec:exp}
The models are experimentally compared on the task of estimating the number of degrees of freedom of five dynamical systems. The latent variables of the model correspond to the state variables of the dynamical system. We comment on the interpretability of the identified state variables for some dynamical systems. 

The data comprises video frames of the systems which are either real, filmed with a camera, or numerically simulated from Chen et al. They comprise 100,000 frames per dynamical system, comprising 100 frames from 1,000 random trajectories. These datasets are divided into training, validation, and test sets, constituting approximately 80\%, 10\%, and 10\% of the data, respectively. 

\begin{figure}[h]
    \centering
    \includegraphics[width=.09\textwidth]{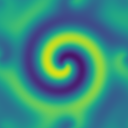}\hfill
    \includegraphics[width=.09\textwidth]{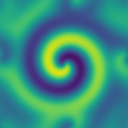}\hfill
    \includegraphics[width=.09\textwidth]{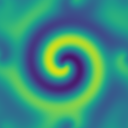}\hfill
    \includegraphics[width=.09\textwidth]{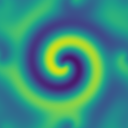}\hfill
    \includegraphics[width=.09\textwidth]{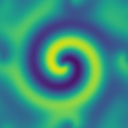}\hfill
    \includegraphics[width=.09\textwidth]{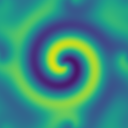}\hfill
    \includegraphics[width=.09\textwidth]{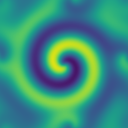}\hfill
    \includegraphics[width=.09\textwidth]{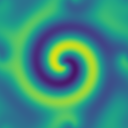}\hfill
    \includegraphics[width=.09\textwidth]{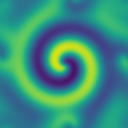}\hfill
    \includegraphics[width=.09\textwidth]{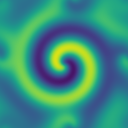}\hfill
    \includegraphics[width=.09\textwidth]{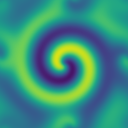}\hfill
    \includegraphics[width=.09\textwidth]{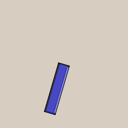}\hfill
    \includegraphics[width=.09\textwidth]{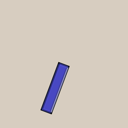}\hfill
    \includegraphics[width=.09\textwidth]{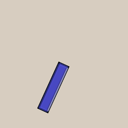}\hfill
    \includegraphics[width=.09\textwidth]{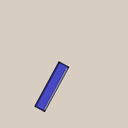}\hfill
    \includegraphics[width=.09\textwidth]{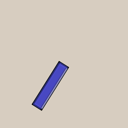}\hfill
    \includegraphics[width=.09\textwidth]{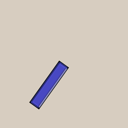}\hfill
    \includegraphics[width=.09\textwidth]{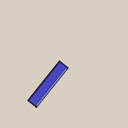}\hfill
    \includegraphics[width=.09\textwidth]{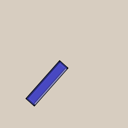}\hfill
    \includegraphics[width=.09\textwidth]{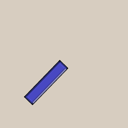}\hfill
    \includegraphics[width=.09\textwidth]{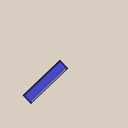}\hfill
    \includegraphics[width=.09\textwidth]{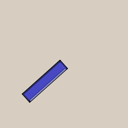}\hfill
    \includegraphics[width=.09\textwidth]{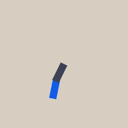}\hfill
    \includegraphics[width=.09\textwidth]{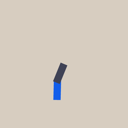}\hfill
    \includegraphics[width=.09\textwidth]{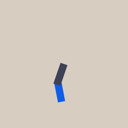}\hfill
    \includegraphics[width=.09\textwidth]{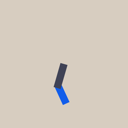}\hfill
    \includegraphics[width=.09\textwidth]{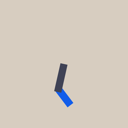}\hfill
    \includegraphics[width=.09\textwidth]{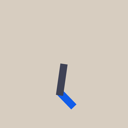}\hfill
    \includegraphics[width=.09\textwidth]{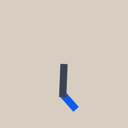}\hfill
    \includegraphics[width=.09\textwidth]{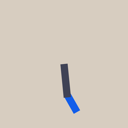}\hfill
    \includegraphics[width=.09\textwidth]{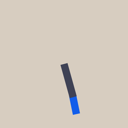}\hfill
    \includegraphics[width=.09\textwidth]{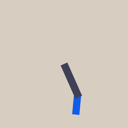}\hfill
    \includegraphics[width=.09\textwidth]{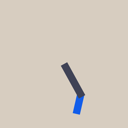}\hfill
    \includegraphics[width=.09\textwidth]{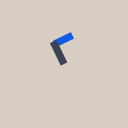}\hfill
    \includegraphics[width=.09\textwidth]{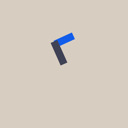}\hfill
    \includegraphics[width=.09\textwidth]{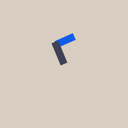}\hfill
    \includegraphics[width=.09\textwidth]{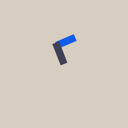}\hfill
    \includegraphics[width=.09\textwidth]{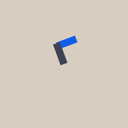}\hfill
    \includegraphics[width=.09\textwidth]{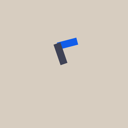}\hfill
    \includegraphics[width=.09\textwidth]{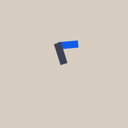}\hfill
    \includegraphics[width=.09\textwidth]{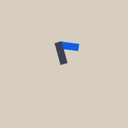}\hfill
    \includegraphics[width=.09\textwidth]{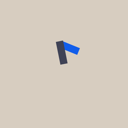}\hfill
    \includegraphics[width=.09\textwidth]{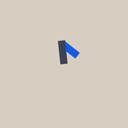}\hfill
    \includegraphics[width=.09\textwidth]{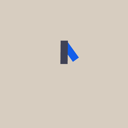}\hfill
    \includegraphics[width=.09\textwidth]{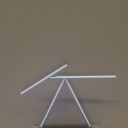}\hfill
    \includegraphics[width=.09\textwidth]{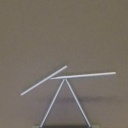}\hfill
    \includegraphics[width=.09\textwidth]{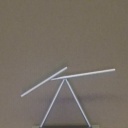}\hfill
    \includegraphics[width=.09\textwidth]{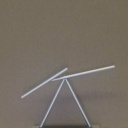}\hfill
    \includegraphics[width=.09\textwidth]{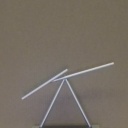}\hfill
    \includegraphics[width=.09\textwidth]{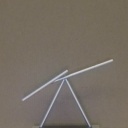}\hfill
    \includegraphics[width=.09\textwidth]{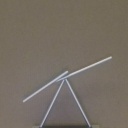}\hfill
    \includegraphics[width=.09\textwidth]{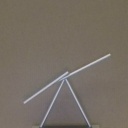}\hfill
    \includegraphics[width=.09\textwidth]{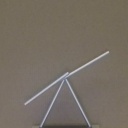}\hfill
    \includegraphics[width=.09\textwidth]{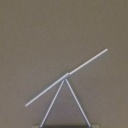}\hfill
    \includegraphics[width=.09\textwidth]{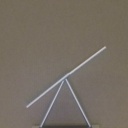}\hfill
    \caption{Sequences of frames of the five dynamical systems: from the top in each row, the reaction-diffusion system, single pendulum system, double pendulum system, elastic pendulum system, and swing stick system from Chen et al.}
    \label{fig:frames}
\end{figure}

The five dynamical systems\footnote{These systems are used as is from the experiments of Chen et al. and are available for download here: https://github.com/BoyuanChen/neural-state-variables} are the reaction-diffusion system (simulated dynamics of a planar spiral wave), a single pendulum system, a double pendulum system, a swing stick system (articulated arms fixed on a rigid base), and an elastic pendulum system (double pendulum which main arm length can vary). Example frames of the five systems are shown in Figure~\ref{fig:frames}. %The former four are real systems filmed using a video camera. % that of a reaction-diffusion system (simulated dynamics of a planar spiral wave) with two state variables, a single pendulum system with two state variables, a double pendulum system with four state variables, a swing stick system (articulated arms fixed on a rigid base) with four state variables and an elastic pendulum system (double pendulum which main arm length can vary) with six state variables. The former four are real systems filmed using a video camera. Example frames of the five systems are shown in Figure~\ref{fig:frames}.

The outermost autoencoder shared by all models takes as input two consecutive frames and outputs the next two consecutive frames. has five consecutive convolutional layers and a latent space of size $64$. The innermost autoencoder takes as input the latent space of the outermost autoencoder and reconstructs it as the output. It has four fully connected layers. The baseline and proposed PI-AE have a latent space of dimension ID. The proposed PI-VAE and HPI-VAE have a latent space of size $10$ which exceeds the ground truth intrinsic dimension of the five dynamical systems. The $\beta$ hyperparameter which adjusts the respective weights of the reconstruction loss and KL divergence is placed on the reconstruction loss term. 
Table~\ref{tab:hyperparams} reports the value of $\beta$ for the PI-VAE and HPI-VAE for the five dynamical systems. All other loss weights for all models are set to 1. All innermost autoencoders are trained for 1000 epochs and have reconstruction losses of less than $0.01$.  
\begin{table}[h]
    \centering
    \caption{Reconstruction loss weights for PI-VAE and HPI-VAE}
    \begin{tabular}{lc@{\hspace{1em}}c@{\hspace{1em}}c@{}}
    \\
    \toprule
        System                 & PI-VAE    & HPI-VAE \\
         \midrule
        Reaction-diffusion      & 7         & 7 \\
        Single pendulum         & 17        & 20 \\
        Double pendulum         & 30        & 40 \\
        Swingstick system       & 30        & 30 \\
        Elastic pendulum        & 50        & 80 \\ 
        \bottomrule
    \end{tabular}
    \label{tab:hyperparams}
\end{table}

The number of degrees of freedom estimated by the baseline and proposed models for the five dynamical systems is shown in Table~\ref{tab:id}. For the baseline and PI-AE, this is the value given by Chen et al.'s intrinsic dimension estimator. For the PI-VAE and HPI-VAE, these are the number of latent variables that are non-zero. A variance threshold of 0.01, corresponding to 1\% of the targeted variance of the reduced prior, discriminates between the non-zero latent variables of interest and the uninformative latent variables.

\begin{table}[h]
    \centering
    \caption{Latent space dimension of systems from different models}
    \begin{tabular}{lc@{\hspace{1em}}c@{\hspace{1em}}c@{\hspace{1em}}c@{\hspace{1em}}c@{\hspace{1em}}c@{\hspace{1em}}}
    \toprule
    System              & Ground truth & Baseline~\cite{chen2022discovery} & PI-AE & PI-VAE & HPI-VAE  \\
    \midrule
    Reaction-diffusion  & 2 & $2.16\approx 2$& $2.16\approx 2$ & 2 & 2 \\
    Single pendulum     & 2 & $2.05\approx2$ & $2.05\approx2$ & 2 & 2 \\
    Double pendulum     & 4 & $4.71\approx 4$ & $4.71\approx 4$ & 4 & 4 \\
    Swingstick system   & 4 & $4.89\approx 4$ & $4.89\approx 4$ & 4   & 4 \\
    Elastic pendulum    & 6 & $5.34\approx 6$ & $5.34\approx 6$ & 6  & 6 \\
    \bottomrule
  \end{tabular}
    \label{tab:id}
\end{table}

Table~\ref{tab:id} shows ground truth degrees of freedom of each dynamical system and the degrees of freedom determined by the baseline and proposed models. The baseline model determines the degrees of freedom by rounding the value from an intrinsic dimension estimator to a whole number. All models can successfully reconstruct the latent space of the outermost autoencoder with a reconstruction loss lower than $0.01$. 

The baseline model can reconstruct the latent space of the outermost autoencoder when the latent space is reduced to the degrees of freedom determined by the intrinsic dimension estimator. The PI-AE, which places an observational bias on the baseline model, can also reconstruct the latent space despite the additional physics constraint placed on the latent space.  
The PI-VAE and HPI-VAE also reconstruct the latent space of the outermost autoencoder, but without using an intrinsic dimension estimator. This is because the KL divergence term can push the latent variables' distributions to resemble the prior. Therefore variables that do not contribute significantly to reducing the reconstruction loss become redundant and converge to zero. This process inherently minimizes the dimensionality of the latent space by eliminating excess variables, thus determining the degrees of freedom of the dynamical system when reconstructing the latent space of the outermost encoder.

Additionally, we comment on the non-zero latent variables of interest and compare them to the latent variables of the baseline and PI-AE by visualizing the latent variables for a random trajectory of the dynamical system. We focus on the latent variables obtained for the three pendulum systems for which the ground truth is known.

\begin{figure}[h]
    \centering
    \begin{subfigure}{0.49\textwidth}
        \centering
        \includegraphics[width=\linewidth]{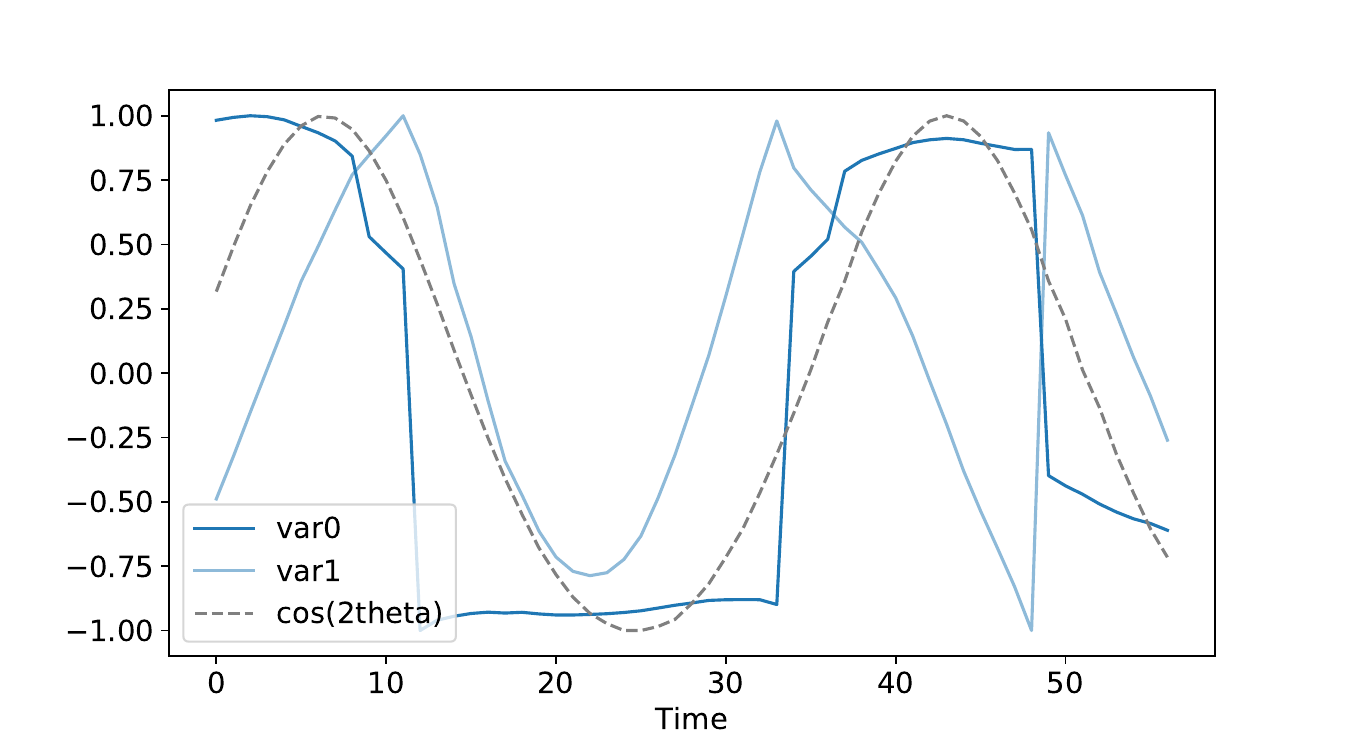}
        \caption{}
        \label{subfigure:singpendbaseline}
    \end{subfigure}
    \begin{subfigure}{0.49\textwidth}
        \centering
        \includegraphics[width=\linewidth]{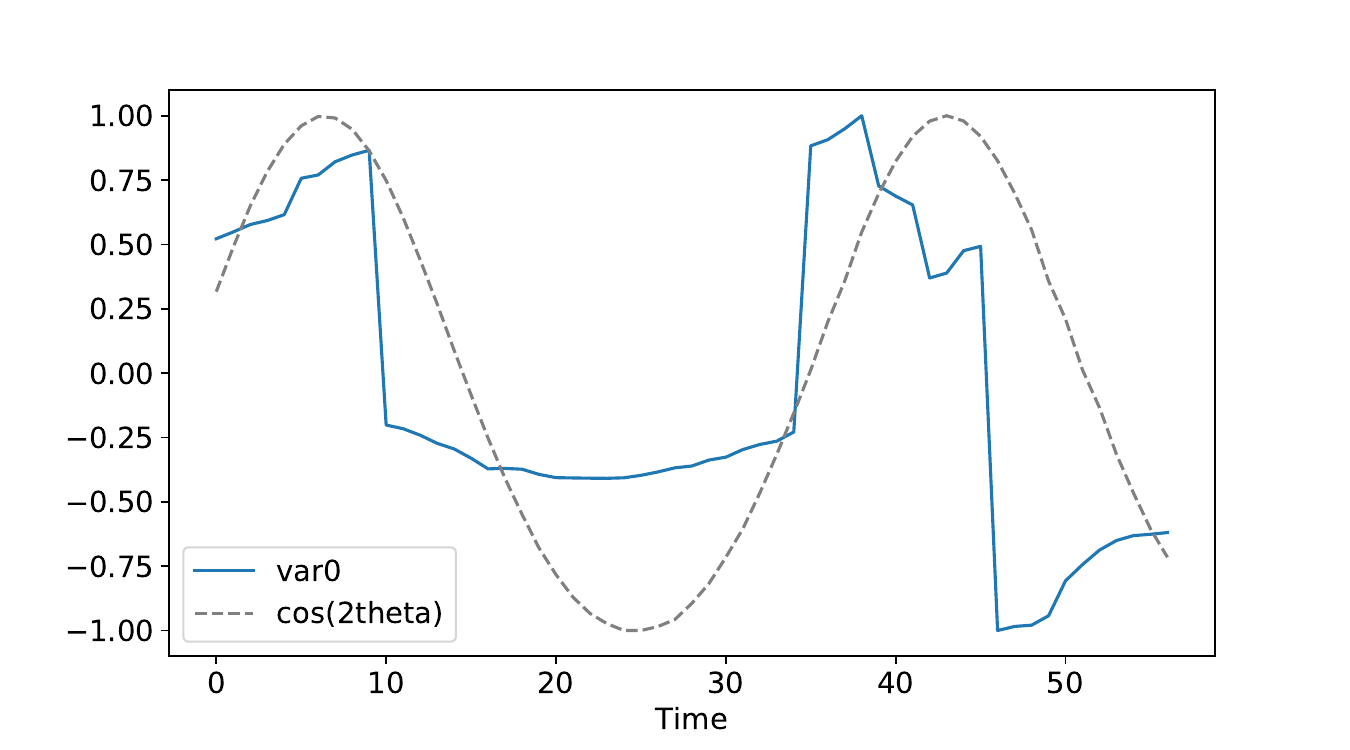}
        \caption{}
        \label{subfigure:vpendpiae}
    \end{subfigure}
    \begin{subfigure}{0.49\textwidth}
        \centering
        \includegraphics[width=\linewidth]{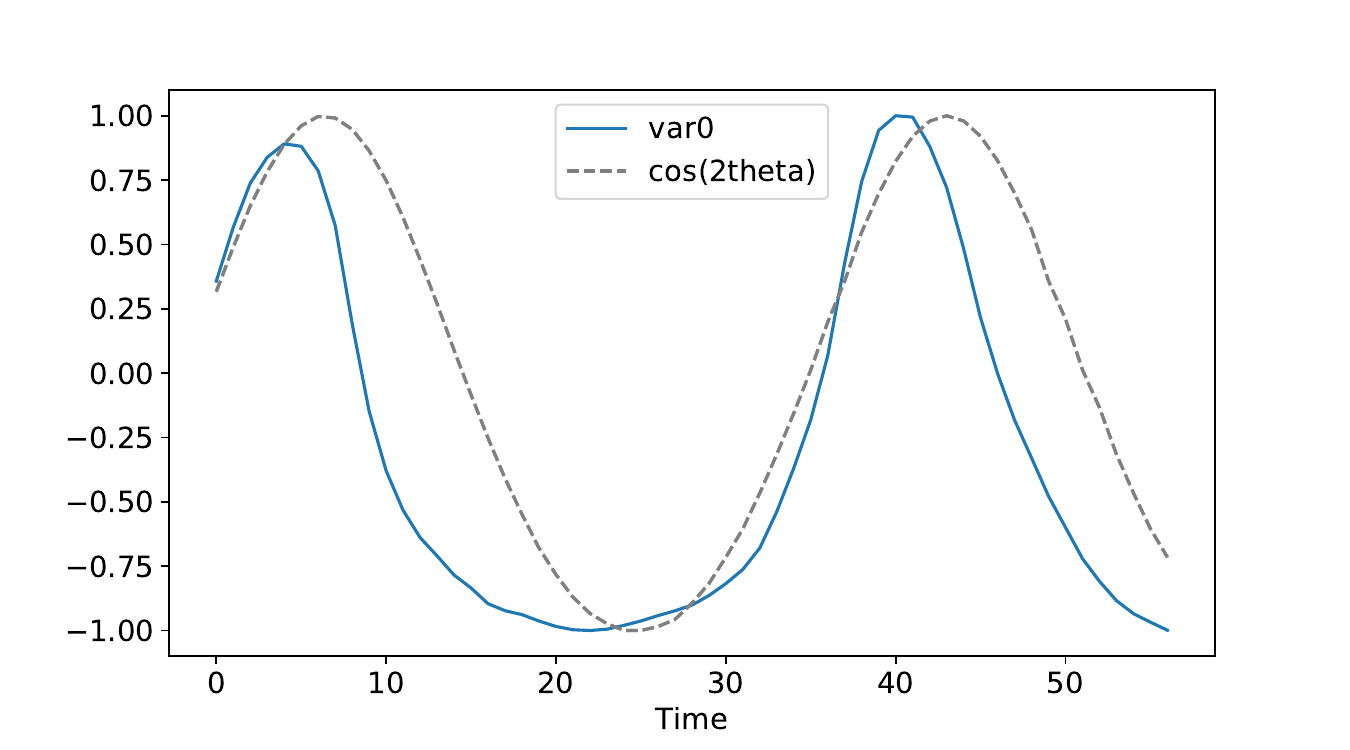}
        \caption{}
        \label{subfigure:singpendpivae}
    \end{subfigure}
        \begin{subfigure}{0.49\textwidth}
        \centering
        \includegraphics[width=\linewidth]{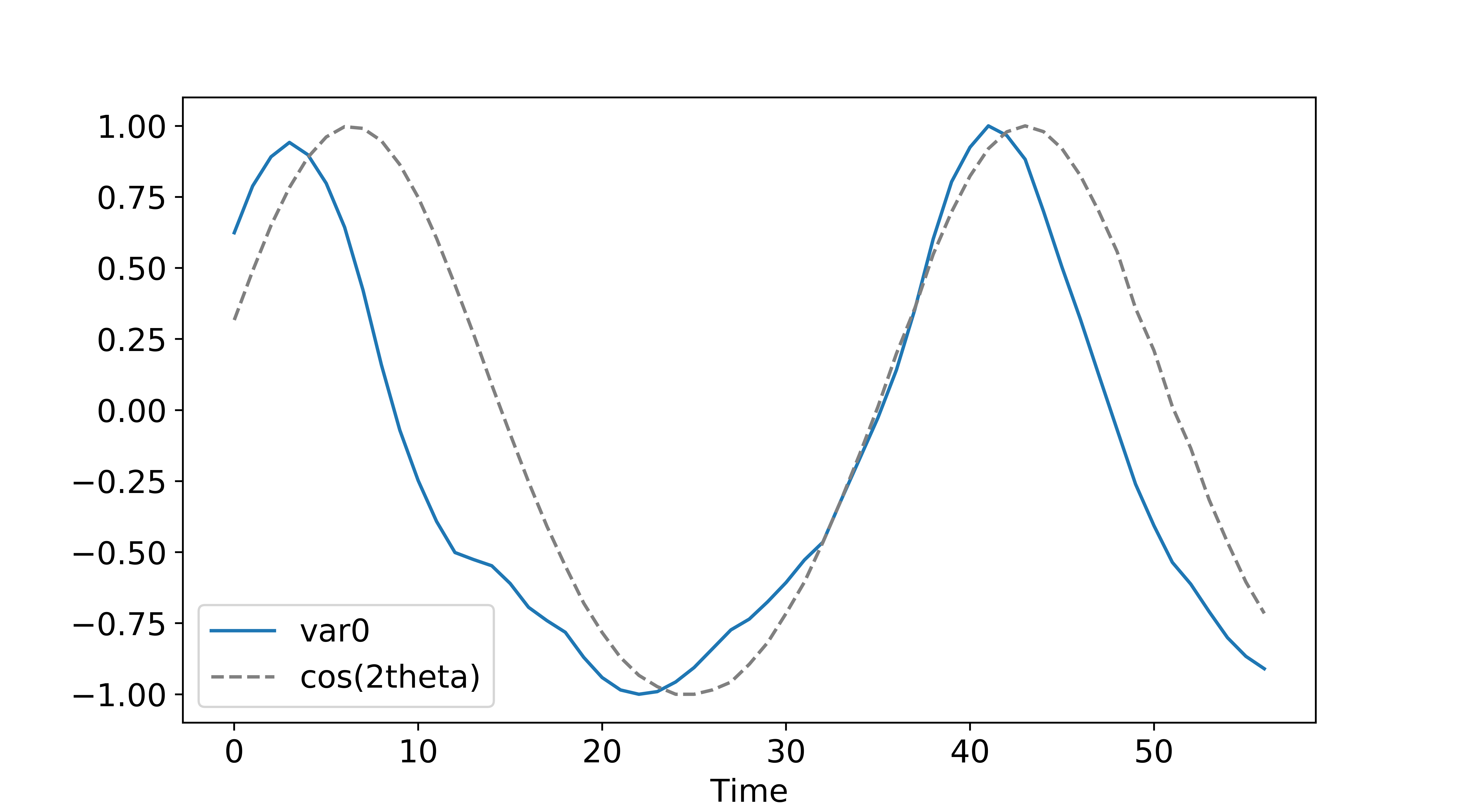}
        \caption{}
        \label{subfigure:singpendhpivae}
    \end{subfigure}
    \caption{Values of the latent variables obtained by the models (y-axis) against time (x-axis) for one trajectory of the simple pendulum. The subplots show the respective latent variables for the (a) baseline, (b) PI-AE, (c) PI-VAE, and (d) HPI-VAE model.}
    \label{figure:vars}
\end{figure}
\textbf{Single pendulum} Figure~\ref{figure:vars} displays the latent variables obtained with each model, scaled to between -1 and +1. The baseline model (a) identifies two latent variables, \verb|var0| and \verb|var1|, which appear entangled or redundant. The PI-AE, PI-VAE, and HPI-VAE capture the system's evolution using two unique latent variables. The first latent variable is shown. The second is its time derivative, omitted for visualization clarity. For all models, the first of the two latent variables are correlated with $\cos{2\theta(t)}$, where $\theta(t)$ is the angle of the pendulum arm with respect to its position of rest at time $t$ given on the x-axis, although this correlation is more apparent with the PI-VAE and HPI-VAE. The plot of $\cos{2\theta(t)}$ against time, $t$, is plotted as a black dotted line on all subplots of Figure~\ref{figure:vars}.
It can also be observed that both the baseline and PI-AE in (a) and (b) exhibit discontinuous changes that align with the curve of $\cos{2\theta(t)}$. However, in the PI-VAE and HPI-VAE models, the latent variable is smoother, revealing a more continuous and coherent understanding of the pendulum's dynamics. In particular, the latent variable of the HPI-VAE has sharper peaks, which are more similar to the peaks of the $\cos{2\theta(t)}$ curve. 

\begin{figure}[h]
    \centering
    \begin{subfigure}{0.49\textwidth}
        \centering
        \includegraphics[width=\linewidth]{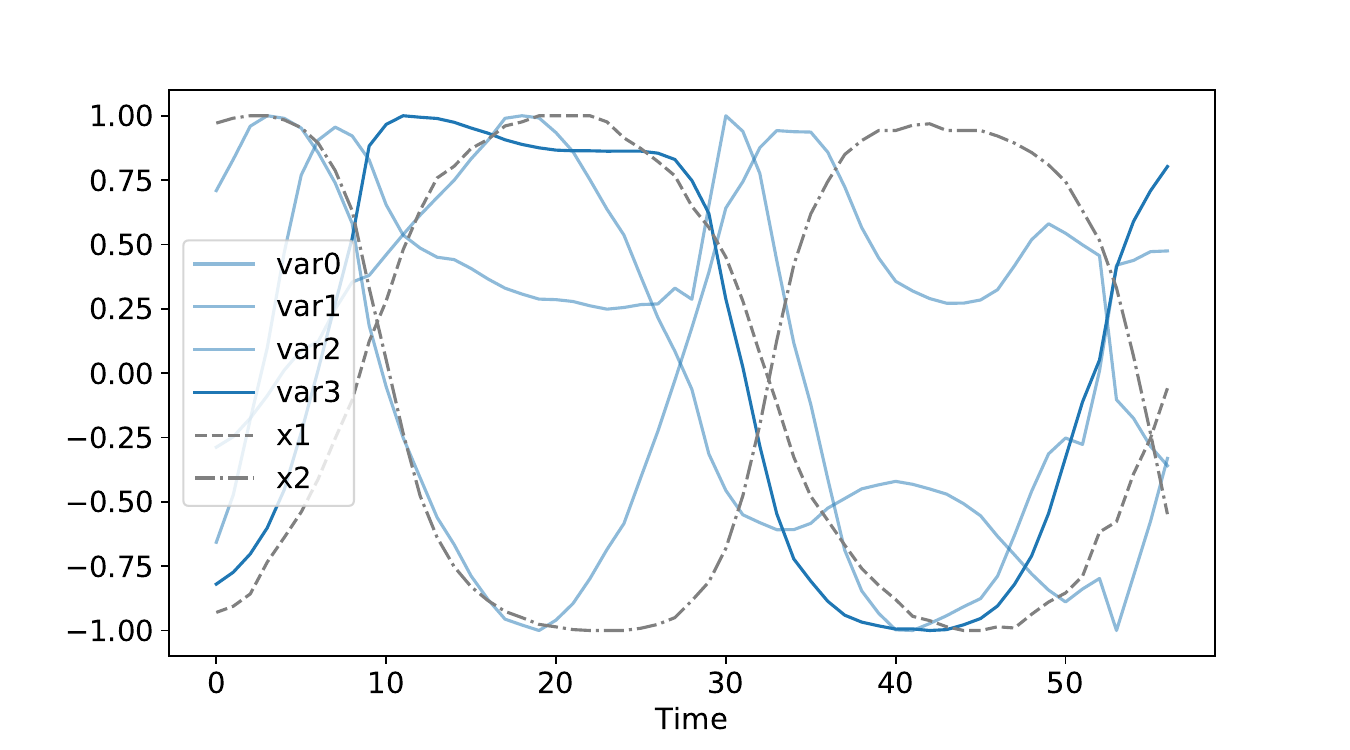}
        \caption{}
        \label{subfigure:doubpendbaseline}
    \end{subfigure}
    \begin{subfigure}{0.49\textwidth}
        \centering
        \includegraphics[width=\linewidth]{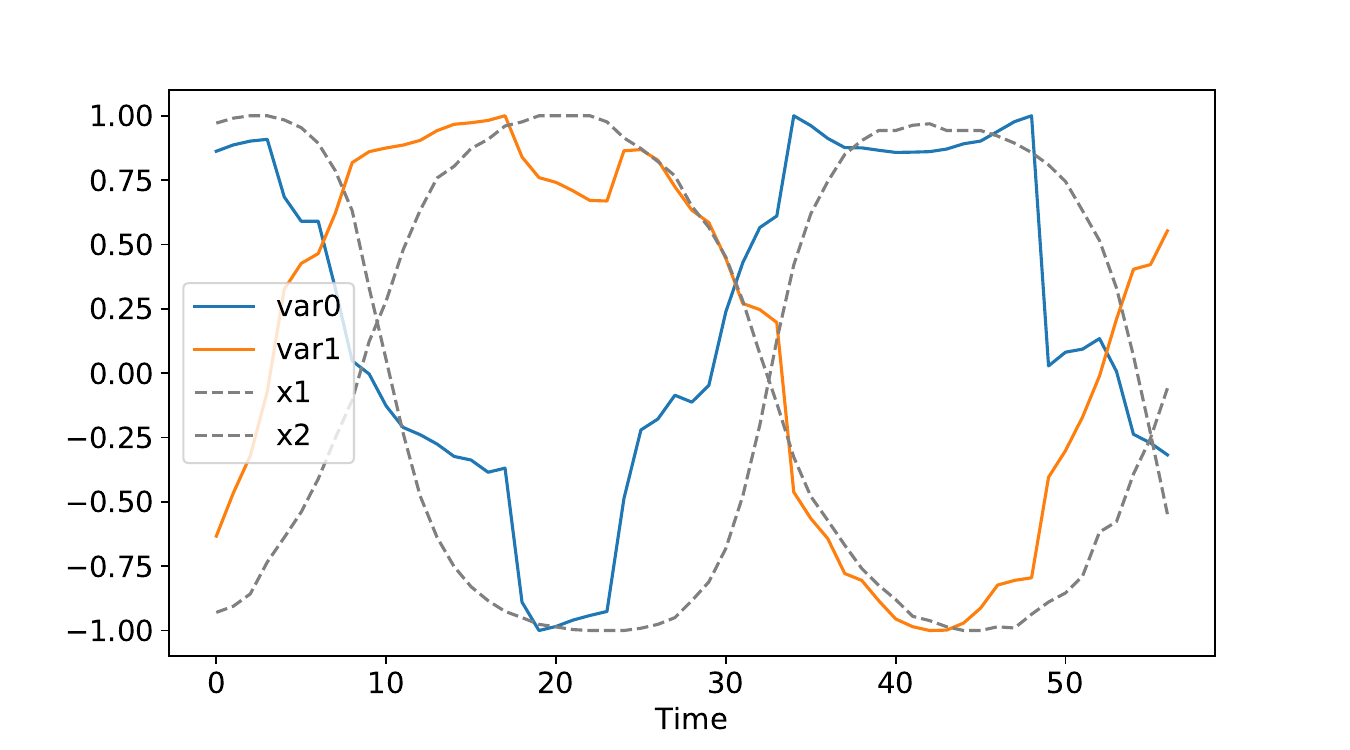}
        \caption{}
        \label{subfigure:doubpendpiae}
    \end{subfigure} \\
    \begin{subfigure}{0.49\textwidth}
        \centering
        \includegraphics[width=\linewidth]{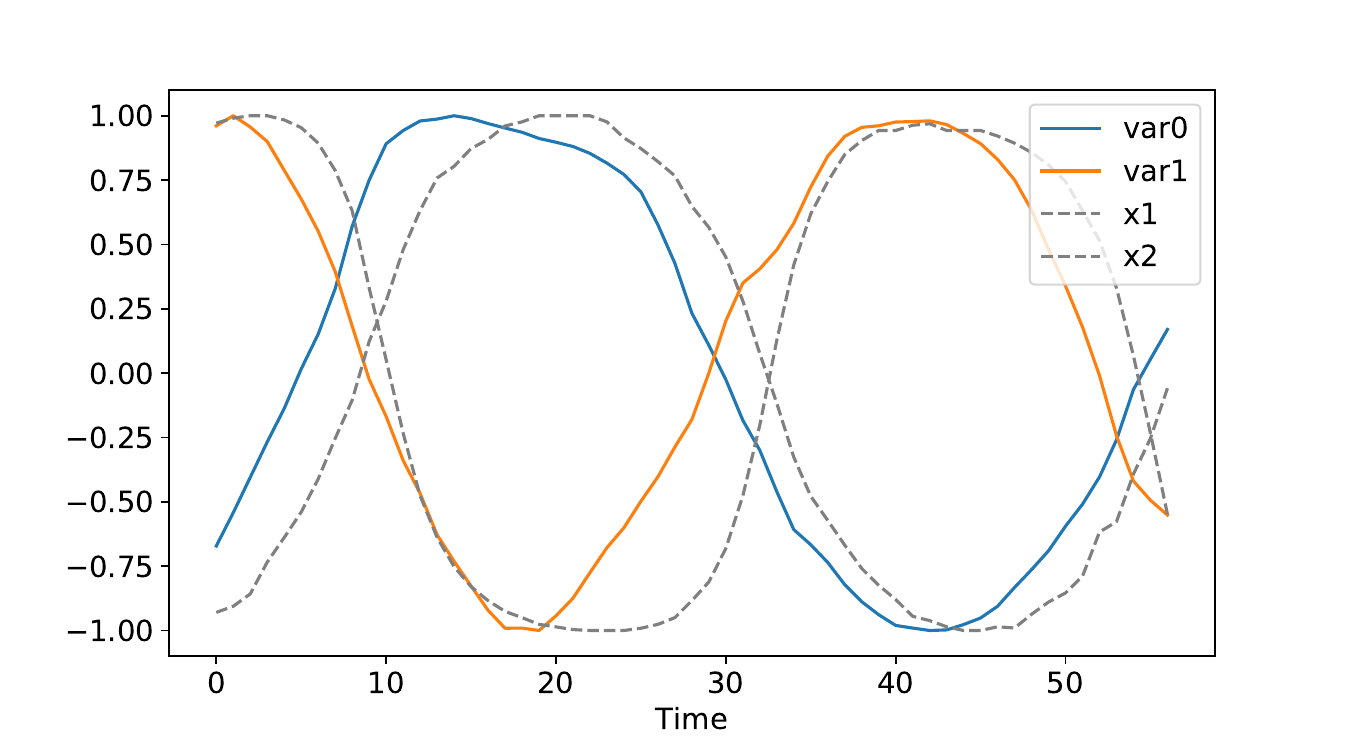}
        \caption{}
        \label{subfigure:doubpendpivae}
    \end{subfigure}
        \begin{subfigure}{0.49\textwidth}
        \centering
        \includegraphics[width=\linewidth]{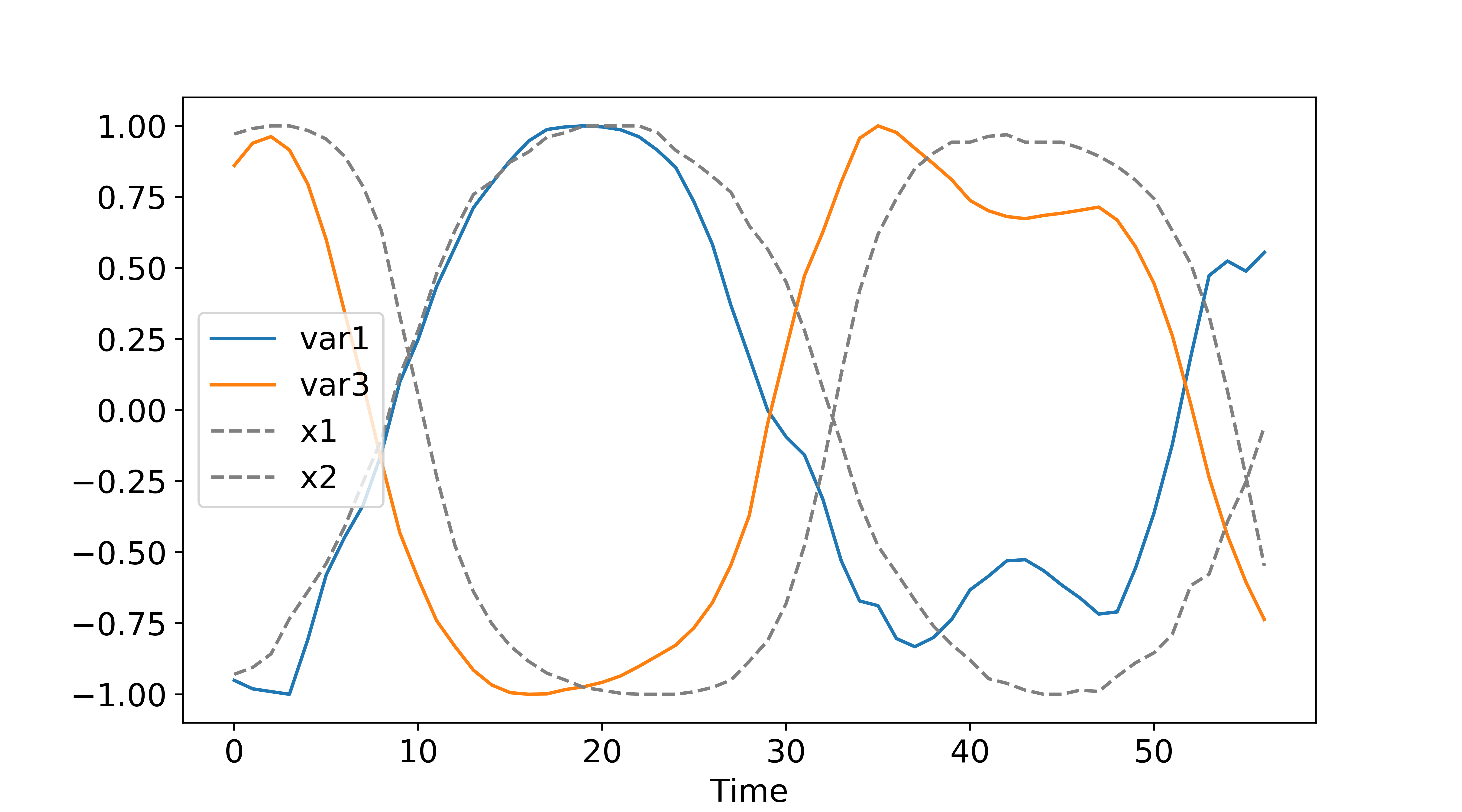}
        \caption{}
        \label{subfigure:doubpendhpivae}
    \end{subfigure}
    \caption{Values of the latent variables obtained by the models (y-axis) against time (x-axis) for one trajectory of the double pendulum. The subplots show the respective latent variables for the (a) baseline, (b) PI-AE, (c) PI-VAE, and (d) HPI-VAE model.}
    \label{figure:varsdp}
\end{figure}
\textbf{Double pendulum} 
Figure~\ref{figure:varsdp} displays the latent variables obtained with each model, scaled to between -1 and +1. The baseline model (a) identifies four latent variables, \verb|var0|, \verb|var1|, \verb|var2|, and  \verb|var3|, which appear entangled or redundant. The PI-AE, PI-VAE, and HPI-VAE capture the system's evolution using four unique latent variables. The first two latent variables are shown. The third and fourth are the time derivatives of the first two, omitted for visualization clarity. The horizontal positions of the two arms of the double pendulum, $x_1(t)$ and $x_2(t)$, are plotted in black dotted lines all subplots of Figure~\ref{figure:varsdp}. 
For the baseline model, the third and fourth latent variables are correlated to $x_1(t)$ and $x_2(t)$, and the first two latent variables are correlated with the third. For the PI-AE, PI-VAE, and HPI-VAE, the first two latent variables are correlated with $x_1(t)$ and $x_2(t)$. The PI-AE demonstrates the ability to distinguish and more accurately represent the joint evolution of both arms. This representation is further enhanced in the PI-VAE and HPI-VAE models, which offer an even more refined and smooth portrayal of the dynamics of the two arms in the double pendulum system.

\begin{figure}[h]
    \centering
    \begin{subfigure}{0.49\textwidth}
        \centering
        \includegraphics[width=\linewidth]{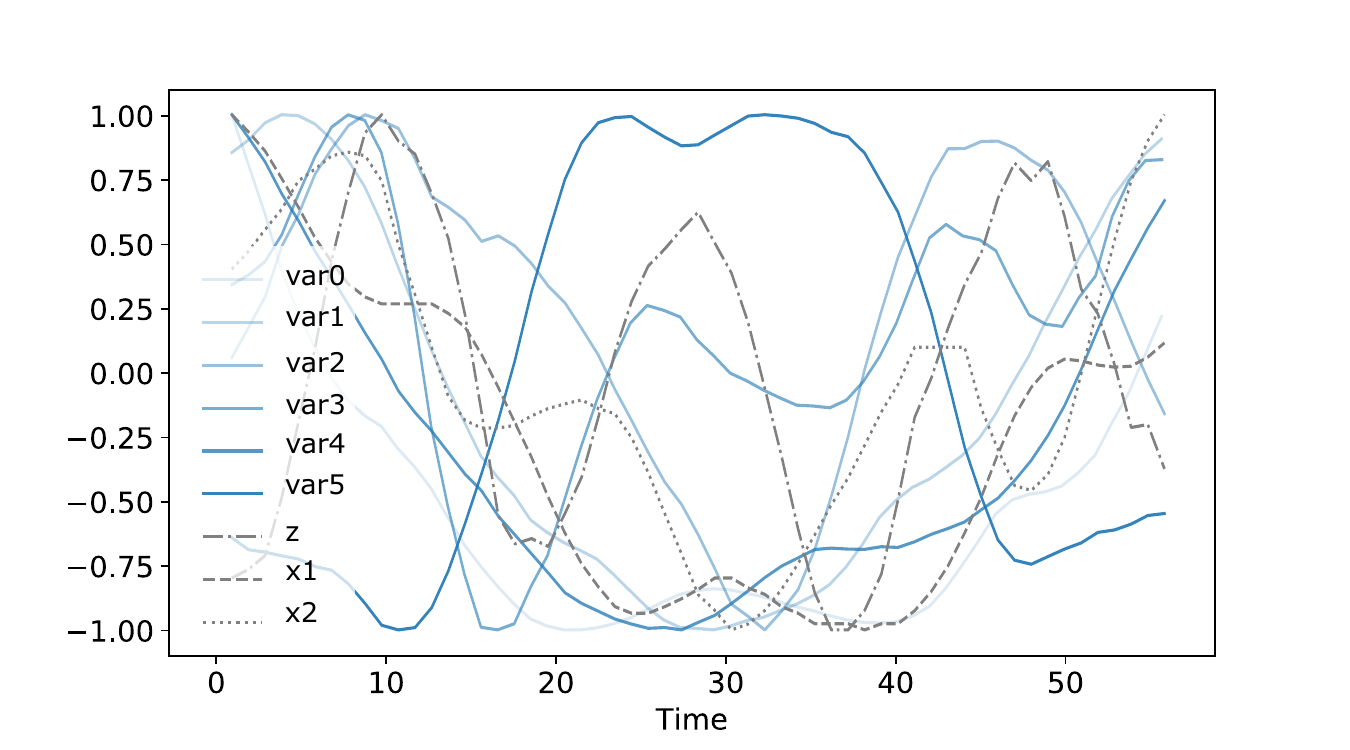}
        \caption{}
        \label{subfigure:elastpendpiae}
    \end{subfigure}
    \begin{subfigure}{0.49\textwidth}
        \centering
        \includegraphics[width=\linewidth]{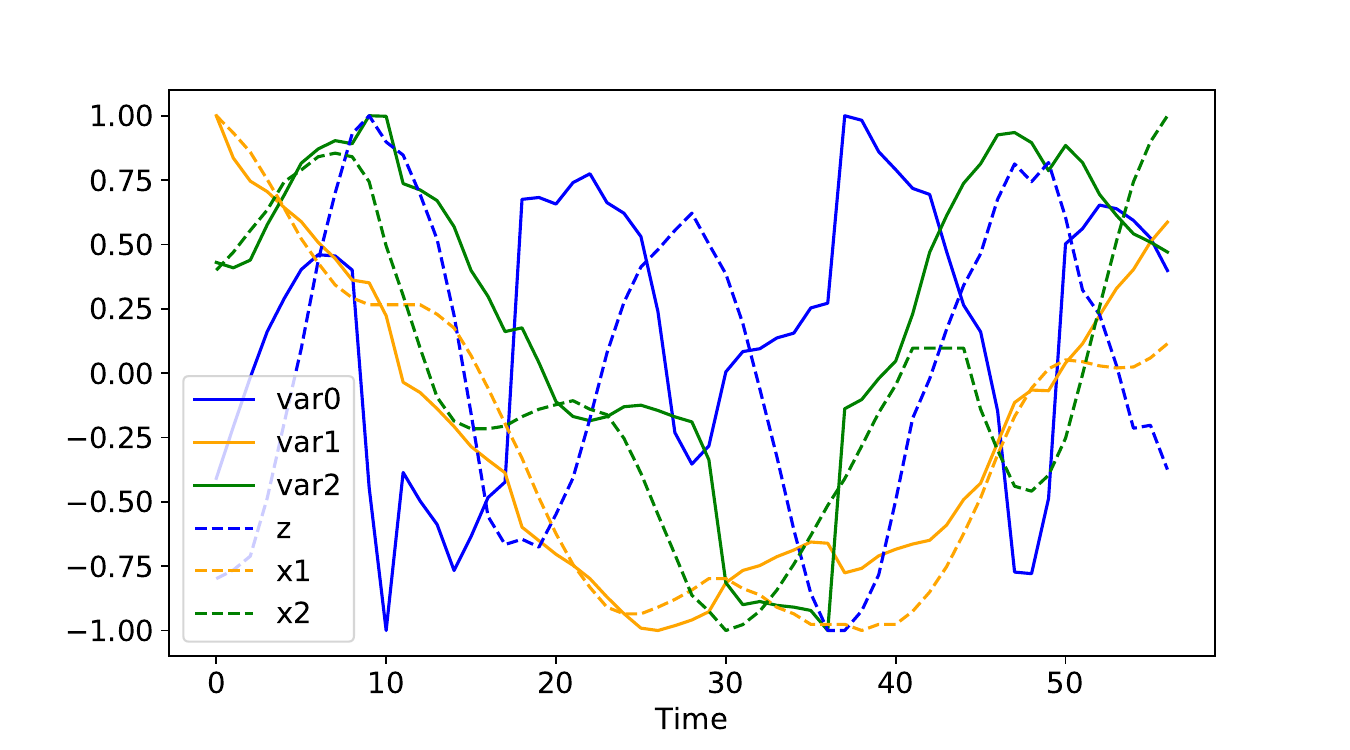}
        \caption{}
        \label{subfigure:elastpendpiae}
    \end{subfigure} \\
    \begin{subfigure}{0.49\textwidth}
        \centering
        \includegraphics[width=\linewidth]{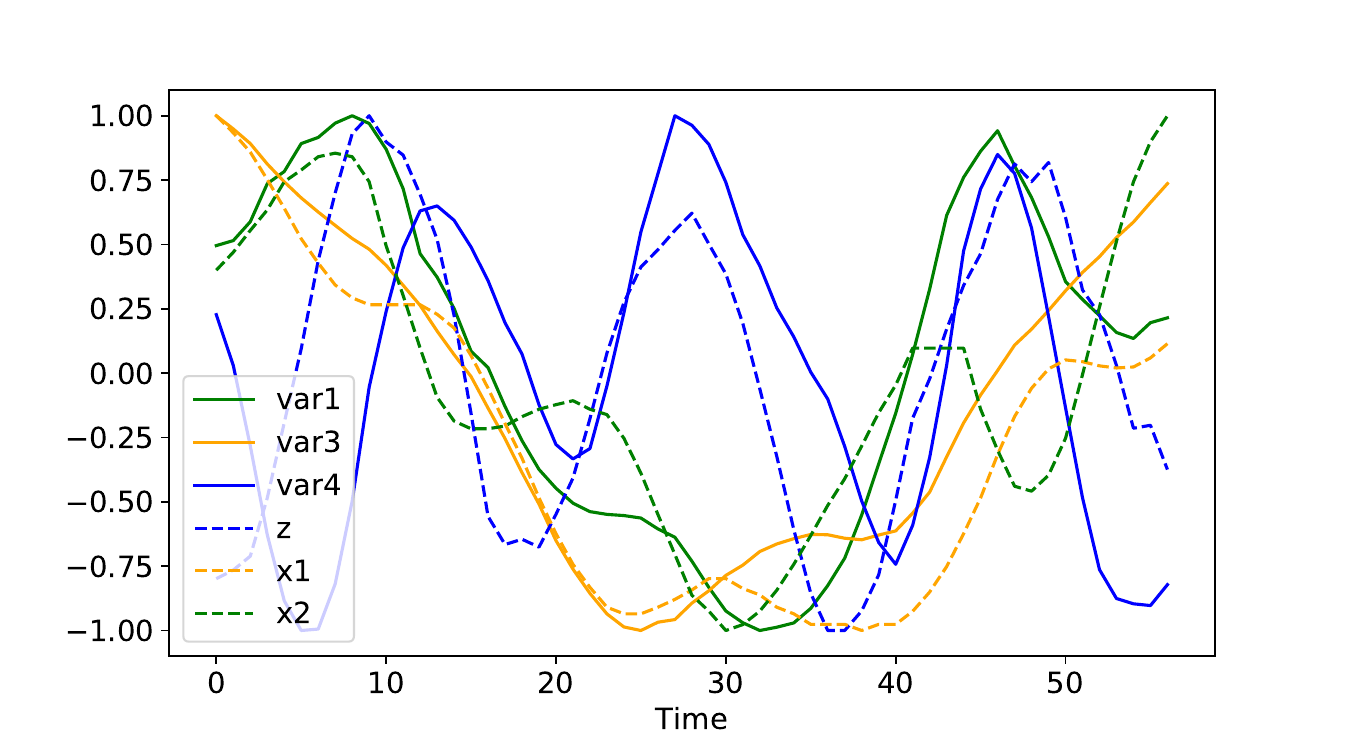}
        \caption{}
        \label{subfigure:elastpendpivae}
    \end{subfigure}
        \begin{subfigure}{0.49\textwidth}
        \centering
        \includegraphics[width=\linewidth]{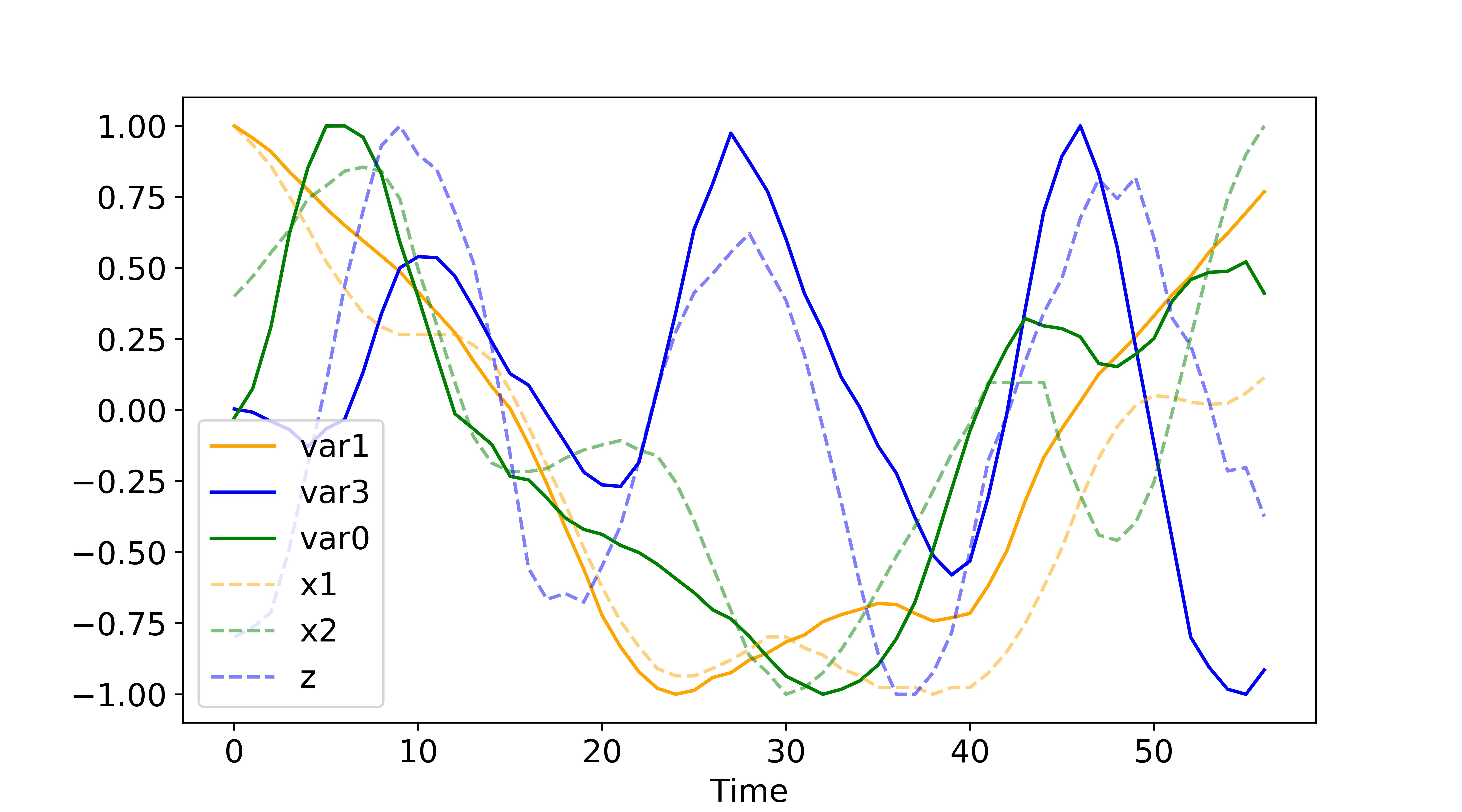}
        \caption{}
        \label{subfigure:elastpendhpivae}
    \end{subfigure}
    \caption{Values of the latent variables obtained by the models (y-axis) against time (x-axis) for one trajectory of the elastic pendulum. The subplots show the respective latent variables for the (a) baseline, (b) PI-AE, (c) PI-VAE, and (d) HPI-VAE model.}
    \label{figure:varsep}
\end{figure}
\textbf{Elastic pendulum} 
Figure~\ref{figure:varsep} displays the latent variables obtained with each model, scaled to between -1 and +1. The baseline model (a) identifies six latent variables, \verb|var0|, \verb|var1|, \verb|var2|,  \verb|var3|, \verb|var4|, and \verb|var5|. The PI-AE, PI-VAE, and HPI-VAE capture the system's evolution using six unique latent variables. The first three latent variables are shown. The last three are the time derivatives of the first three, omitted for visualization clarity. The horizontal positions of the two arms of the elastic pendulum, $x_1(t)$ and $x_2(t)$, and the changing length of its main arm, $z$, are plotted in black dotted lines all subplots of Figure~\ref{figure:varsdp}. 
For the baseline model, the fifth latent variable is correlated with $x_1(t)$. It is not possible to identify the correlation of the other variables with $z$ and $x_2$. Similarly, the latent variables of the PI-AE also do not correlate with $z$ and $x_2$, as seen in Figure~\ref{figure:varsep} (b). However, the PI-VAE in Figure~\ref{figure:varsep} (c) and HPI-VAE in Figure~\ref{figure:varsep} (d) can assign latent variables to model each arm's position and the main arm's fluctuating length. Although the match between the physical and model-derived variables in PI-VAE and HPI-VAE models is not as precise as in simpler cases, these models manage to trace the physical variables' patterns more effectively.

% \begin{figure}[h!]
%     \centering
%     \begin{subfigure}{0.49\textwidth}
%         \centering
%         \includegraphics[width=\linewidth]{plots/vars_models/elastic_pendulum_pivae_all.pdf}
%         \caption{PI-VAE}
% \label{subfigure:elastpendbaseline}
%     \end{subfigure}
%     \begin{subfigure}{0.49\textwidth}
%         \centering
%         \includegraphics[width=\linewidth]{plots/vars_models/elastic-pendulum-HPI-VAE.png}
%         \caption{HPI-VAE}
%         \label{subfigure:elastpendbaseline}
%     \end{subfigure}
%     \caption{Latent variables for the elastic pendulum obtained compared to the main arm position $x_1$, secondary arm position $x_2$, and  variable arm length $z$ using the (a) PI-VAE model and (b) HPI-VAE model.}
%     \label{figure:varsep}
% \end{figure}

%We now want to quantitatively assess the accuracy of the fitting procedure within these two (PI-VAE and HPI-VAE) model frameworks. The correlation coefficients for regressions conducted across all test set trajectories are presented in Table~\ref{tab:corrcoef}. For each physical variable, a distinct regression is fitted and evaluated. The fitting functions employed are those highlighted in the graphical representations; specifically, \( f(\theta) = \cos(2\theta) \) for the single pendulum, \( f(\theta_i) = x_i, i\in\{1,2\} \) for the double and elastic pendulums, and \( f(z) = z \) for the elastic pendulum.

In summary, we have improved the baseline model by addressing its oversight of physics constraints. The proposed models can accurately identify the degrees of freedom of each dynamical system. An additional benefit is that the state variables are more interpretable. 
This was achieved by integrating properties of the examined dynamical systems into the baseline model using physics-informed machine learning. 

The first physics constraint is integrated in the form of an observational bias which separates the spatial and temporal contributions of the obtained state variables. %This capability is effective for all systems, and highlighted in the analysis of pendulum systems, Figure~\ref{figure:vars}(b), ~\ref{figure:varsdp}(b), and ~\ref{figure:varsep}(a). 
However, the method still relies on the estimator of intrinsic dimension, an external, purely statistical tool that overlooks the physics of the system under study. Furthermore, the latent variables are uninterpretable.
The second physics constraint is integrated in the form of a learning bias using a variational autoencoder architecture in the PI-VAE model. It allows for the identification of the number of degrees of freedom for all dynamical systems, as shown in Table~\ref{tab:id}. This enhancement effectively isolates the system's main sources of variation to identify its state variables. An additional benefit is that it yields a structured, continuous latent space. %, as illustrated in the pendulums examples Figure~\ref{figure:vars}(c), ~\ref{figure:varsdp}(c), ~\ref{figure:varsep}(b). However%, as shown in the regression analysis, Table~\ref{tab:corrcoef}, the method seems limited in predicting low-amplitude variations, such as the variable part of the elastic pendulum, which is overshadowed by the larger variations of other system components. A 
The third physics constraint is an inductive bias, which leverages the conservation of the Hamiltonian in systems. This constraint tweaks the interpretability of the state variables, and does not impact its ability to find the degrees of freedom of the dynamical system. % does not appear to significantly impact the metrics studied. However, we believe this feature, without adding too much complexity to the model, can enhance future analyses by enabling energy-conservation insights, predictive accuracy, and robustness against perturbations.

In conclusion, the PI-VAE and HPI-VAE models are capable of identifying the degrees of freedom of various dynamical systems and faithfully modeling the state variables of the various dynamical systems such that they are more interpretable.

\section{Conclusion}

In conclusion, through the integration of physics-informed machine learning with variational autoencoders, we have allied physical knowledge and data-driven machine learning to enhance the interpretability and simplicity of modeling complex systems. Our approach marks an improvement over traditional methods by parsimoniously identifying the degrees of freedom of dynamical systems. Furthermore, the latent variables are a minimal, non-redundant representation of dynamics that faithfully captures the system's physical characteristics. This advancement holds promise for a wide range of applications, from fundamental physics to engineering. We anticipate that the methodologies and insights gleaned from this work will help catalyze further research, fostering the development of more sophisticated, physics-informed models capable of tackling the complexities inherent in the natural and engineered world. 

Future work should be directed toward the interpretation of latent variables. This can involve extending the proposed models to increasingly complex systems, using other physics constraints, or encompassing different modified loss functions. Different state variables have varying contributions to their respective dynamical systems, being able to differentiate the penalization of errors for different latent variables at large and small spatial or temporal scales will be beneficial to identifying more interpretable state variables. 

\bibliographystyle{splncs04}
\bibliography{mybibliography}

\end{document}